\documentclass{article}
\usepackage{ijcai23}
\usepackage{times}
\usepackage{soul}
\usepackage{url}
\usepackage[hidelinks]{hyperref}
\usepackage[utf8]{inputenc}
\usepackage[small]{caption}
\usepackage{graphicx}
\usepackage{amsmath}
\usepackage{amsthm}
\usepackage{booktabs}
\usepackage{algorithm}
\usepackage{algorithmic}
\usepackage[switch]{lineno}
\usepackage[bb=boondox]{mathalfa}

\newtheorem{property}{Property}
\newtheorem{definition}{Definition}

\usepackage{comment}

\title{Scalable Coupling of Deep Learning with Logical Reasoning}

\author{Marianne Defresne$^{1,2}$\and Sophie Barbe$^2$\And Thomas Schiex$^1$
\affiliations
$^1$Université Fédérale de Toulouse, ANITI, INRAE, UR 875, 31326 {Toulouse}, France\\
$^2$TBI, Université de Toulouse, CNRS, INRAE, INSA, ANITI, 31077 Toulouse, France
\emails
\{marianne.defresne, sophie.barbe\}@insa-toulouse.fr, thomas.schiex@inrae.fr}

\def\by{\mathbf{y}}
\def\bt{\mathbf{t}}
\def\bv{\mathbf{v}}
\def\bY{\mathbf{Y}}
\def\bM{\mathbf{M}}

\DeclareMathOperator*{\argmin}{\arg\!\min}

\begin{document}

\maketitle

\begin{abstract}
  In the ongoing quest for hybridizing discrete reasoning with neural nets, there is an increasing interest in neural architectures that can learn how to solve discrete reasoning or optimization problems from natural inputs. In this paper, we introduce a scalable neural architecture and loss function dedicated to learning the constraints and criteria of NP-hard reasoning problems expressed as discrete Graphical Models. Our loss function solves one of the main limitations of Besag's pseudo-loglikelihood, enabling learning of high energies. We empirically show it is able to efficiently learn how to solve NP-hard reasoning problems from natural inputs as the symbolic, visual or many-solutions Sudoku problems as well as the energy optimization formulation of the protein design problem, providing data efficiency, interpretability, and \textit{a posteriori} control over predictions.
\end{abstract}

\section{Introduction}

In recent years, several hybrid neural architectures have been proposed to integrate discrete reasoning or optimization within neural networks. Many of the architectures incorporate an optimization or reasoning layer in a neural network where the previous layer outputs the parameters defining the criteria of the discrete problem~\cite{satnet,OptNet,mandi2020interior,blackbox,TiasLTR,blackbox2}. Learning is then achieved by back-propagating gradients from known solutions and natural inputs (such as symbols, images, text or molecule geometries). In this paper, we are more specifically interested in scalable learning when the underlying discrete reasoning problem incorporates unknown logical (deterministic) information or constraints.

Our contributions are twofold: we introduce a hybrid architecture combining arbitrary deep learning layers with a final discrete NP-hard Graphical Model (GM) reasoning layer and propose a new loss function that efficiently deals with logical information. We use discrete GMs~\cite{cooper2020graphical} as the reasoning language because GMs have been used to represent both Boolean functions (in propositional logic, constraint networks) and constrained numerical functions (partial weighted MaxSAT, Cost Function Networks). Also, a probabilistic interpretation of such models is available in stochastic GMs such as Markov Random Fields (MRFs), where infinite costs represent zero probabilities (infeasibility).

Besides the discrete nature of variables that defines zero gradients almost everywhere, the Boolean nature of logic creates an additional challenge. Indeed, constraints are either satisfied or violated and the continuous relaxation of Boolean satisfiability or constraint satisfaction in weighted variants~\cite{satnet,brouard2020pushing,RRN} may not always account for the potentially redundant nature of constraints: in some contexts, a meaningful constraint $c$ may be implied by a set of other constraints $C$, making the learning of $c$ from examples impossible if $C$ has already been learned.

In our architecture, during inference, an arbitrary deep learning architecture receives natural inputs and outputs a fully parameterized discrete GM. This GM can then be solved using any suitable optimization solver. To benefit from the guarantees of logical reasoning on proper input, we use the exact GM prover toulbar2~\cite{allouche2015anytime}. During learning, GMs computed from natural inputs need to be gradually improved from solutions in the data set. Given the NP-hard nature of discrete GM reasoning and our target of scalable learning, using an exact optimization during learning seems inadequate: even if the finally predicted GMs may be empirically easy to solve (as is the case, \emph{e.g.}, for the Sudoku problem), the GMs predicted in the early epochs of learning are essentially random, defining empirically extremely hard instances that cannot be solved to optimality in a reasonable time~\cite{zhang2001phase}. Relying instead on more scalable convex relaxations of the discrete GM optimization problem~\cite{durante:hal-03673535} would come at the cost of sacrificing the guarantees of logical reasoning on proper input.

We instead exploit the probabilistic interpretation of weighted GMs and target the optimization of the asymptotically consistent and scalable Negative Pseudo-LogLikelihood (NPLL) of the training data set~\cite{besag1975statistical}. However, this loss function is plagued with an incapability of dealing with large costs~\cite{difficultGM} and therefore with constraints. In this paper, we analyze the reasons for this incapability and show that it can be explained by the context-dependent redundancy of logical information. We propose a variant, the E-NPLL, for scalable decision-focused learning. With this differentiable informative loss, our architecture is able to efficiently learn from natural inputs to, \emph{e.g.}, solve the Visual Sudoku~\cite{satnet,brouard2020pushing}. We also apply it to a many-solutions Sudoku data set~\cite{1ofMany} and to a data set of Protein Design instances including several hundreds of variables. With an exact prover used during inference, 100\% accuracy can be reached on the symbolic Sudoku, even with small data sets. Moreover, the output of the penultimate layer being a full GM, it can be scrutinized to detect possible symmetries~\cite{lim2022learning} for example. It is also possible to introduce \emph{a posteriori} constraints to get solutions that satisfy desirable properties or even bias the learned criteria by adding new functions to the produced GM.

\section{Preliminaries}

\subsection{Background}

A discrete graphical model is a concise description of a joint function of many discrete variables as the combination of many simple functions. Depending on the nature of the output of the functions (Boolean or numerical), and how they are combined and described, GMs cover a large spectrum of AI NP-hard reasoning and optimization frameworks including Constraint Networks, Propositional Logic as well as their numerical additive variants Cost Function Networks and partial weighted MaxSAT~\cite{cooper2020graphical}. Following cost exponentiation and normalization, these numerical joint functions can describe joint probability distributions, as done in Markov Random Fields (MRFs). In this paper, we use Cost Function Networks for their ability to express both numerical and logical functions. We assume here that cost functions take their value in $\bar{\mathbb{R}} = \mathbb{R}\cup \{\infty\}$.

In the rest of the document, we denote sequences, vectors, tensors in bold. Variables are denoted in capitals with a given variable $Y_i \in \bY$ being the $i^{th}$ variable in the sequence $\bY$. An assignment of the variables in $\bY$ is denoted $\by$ and $y_i$ is the assignment of $Y_i$ in $\by$. $\bY_{-i}$ denotes the sequence of variables $\bY$ after removal of variable $Y_i$ and similarly for $\by_{-i}$ given $\by$. For a set of variables $\bM\subset\bY$, we also note $\bY_{-M} = \bY\setminus\bM$ and $\by_{-M}$ their values in $\by$. The domain of a variable $Y_i$ is a set denoted $D^i$ with $|D^i|\leq d$, the maximum domain size. For a sequence of variables $\bY$ we denote as $D^{\bY}$ the Cartesian product of all $D^i$ with $Y_i\in\bY$. A cost function over a subset of $\bY$ described by a tensor (cost matrix) over $\bar{\mathbb{R}}$ is called an \emph{elementary} cost function.

\begin{definition}
  Given a sequence $\bY = \{Y_1, \ldots, Y_n\}$ of $n$ finite domain variables, a cost function network $C$ is defined as a set of elementary cost functions. It defines a joint cost function, also denoted $C(\cdot) = \sum_{F\in C} F$, involving all variables in $\bY$. The optimization problem, known as the Weighted Constraint Satisfaction Problem (WCSP), is to find an assignment $\by$ that minimizes the joint function $C(\by)$. If $C(\by)< \infty$, $\by$ is called a solution (a model in propositional logic).
\end{definition}
In stochastic GMs such as MRFs, the joint function $C(\cdot)$ is used to define a joint probability distribution $P^C~\propto~exp(-C(\cdot))=\prod_{F\in C} exp(-F)$ requiring the expensive computation of a \#-P-hard normalizing constant.

When a given function $F$ is never larger than another function $F' (F\leq F')$, $F$ is known as a relaxation of $F'$. A constraint is a cost function $F$ such that $F(\bt) \in \{0,\infty\}$: it exactly forbids all assignments $\bt$ such that $F(\bt)=\infty$. When $F \leq F'$ are constraints, we say that $F$ is a logical consequence of $F'$: whenever $F'$ is satisfied (equal to $0$), $F$ is satisfied too. For a set of constraints $C$, $F\in C$ is redundant w.r.t.\ $C$ iff $C$ and $C \setminus\{F\}$ define the same function. At a finer grain, we say $F$ is partially redundant if $\exists F' < F$ such that  $(C\setminus\{F\})\cup\{F'\}$ and $C$ define the same function. Consider for example $\bY=\{Y_1,Y_2,Y_3,Y_4\}$ with domains $\{0,1\}$ and $C=\{Y_1\neq Y_2, Y_2 + Y_3 > 1 ,Y_3\neq Y_4\}$. No constraint is redundant in $C$, but in the context of the assignment $\{Y_2=1,Y_3=1\}$,  the constraint $Y_2 + Y_3 > 1$ becomes redundant w.r.t. $C'=C\cup\{Y_2=1,Y_3=1\}$ . In the context of $\{Y_1=0\}$, $Y_2 + Y_3> 1$ becomes partially redundant, as it could equivalently be replaced by the weaker $Y_2 = Y_3$. Because of redundancies, the observed values in a sample can create a context that makes some constraints redundant and therefore not learnable.

For $n$ variables, a strictly pairwise graphical model $C$ ($\forall F \in C$, $F$ involves exactly $2$ variables) can be described with $n(n-1)/2$ elementary cost function with tensors (matrices) of size at most $d^2$. We denote by $C[i,j]$ the tensor describing the cost function $F_{ij}$ between variables $Y_i$ and $Y_j$. Thus $F_{ij}(Y_i=a,Y_j=b)$ is simply $C[i,j](a,b)$.

\subsection{Problem Statement}

\begin{figure*}[tbh]
  \centering
  \includegraphics[width=0.9\textwidth]{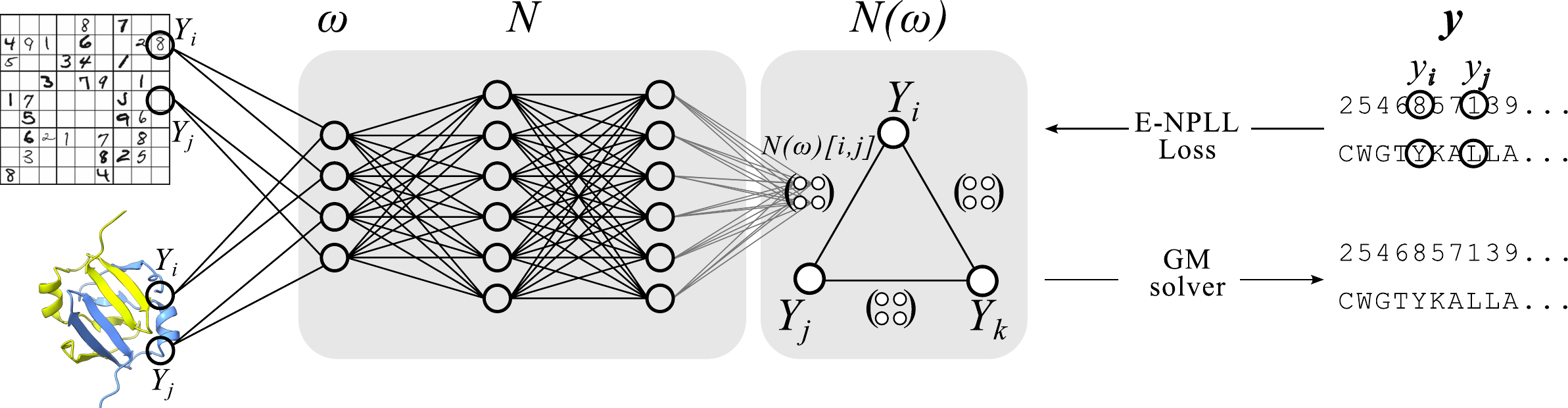}
  \caption{Our hybrid learning architecture: natural inputs $\omega$ (left) feed a neural net $N$ in charge of predicting all pairwise cost functions $F_{ij}$ of the GM $N(\omega)$. To learn $N$, we back-propagate solutions $\by\in S$ through the E-NPLL loss function. At inference, $N(\omega)$ can be directly fed to any GM solver, be it exact, based on a scalable relaxation or a (meta)-heuristics. This is illustrated here on 2 possible problems: a visual Sudoku problem (top) and a protein design problem (bottom).}
  \label{fig:pipeline}
\end{figure*}

In this work, we assume that we observe samples $(\omega,\by)$ of the values $\by$ of the variables $\bY$ as low-cost solutions of an underlying constrained optimization problem with parameters influenced by natural inputs $\omega$. From a data set $S$ of pairs $(\omega, \by)$, we want to learn a model $N$ (in our case, a neural network) which predicts a pairwise graphical model $N(\omega)$ such that $\by\in \argmin_{\by\in D^{\bY}} N(\omega)(\by)$. Such a graphical model $C = N(\omega)$ defines the last layer of our hybrid neural+graphical model architecture (see Figure~\ref{fig:pipeline}).

In terms of supervision, we assume that the variables in $\bY$ are identified but we also want to exploit any information that would be available on elements of $\omega$. Some of these natural inputs may be direct constraints or assignments of variables in $\bY$ that can be directly incorporated into the GM $N(\omega)$, others may be known to influence only a subset of all the variables $\bY$. In the symbolic Sudoku problem, a partially filled grid of numbers is observed in $\omega$ and each observed value in the grid is known to be the value of its corresponding variable. Similarly, in the visual Sudoku, each image of a digit in the grid is known to represent the value of a single variable, observed in $\by$, providing grounding information~\cite{assessingSATNet}.

Assuming the data set $S$ contains i.i.d.\ samples from an unknown probability distribution $P(\bY|\omega)$, a natural loss function for the GM $N(\omega)$ is the asymptotically consistent negative logarithm of the probability $\textrm{NLL}(S) = -\log(\prod_{(\omega,\by)\in S} P^{N(\omega)}(\bY=\by))$ of the observed samples. This negative loglikelihood is however intractable because of the \#P-hard normalizing constant. We instead rely on the asymptotically consistent tractable negative pseudo-loglikelihood~\cite{besag1975statistical} $\textrm{NPLL}(S) = \sum_{(\omega,y)\in S} -\log(\prod_i P^{N(\omega)}(y_i|\by_{-i}))$. The NPLL works at the level of each variable $Y_i$, in the context of $\by_{-i}$, the assignment of all other variables. It  requires only normalization over one variable $Y_i$, a computationally easy task.

\begin{property}
  For $Y_i$, we have $P(Y_i|\by_{-i}) = \mathrm{softmax}(-m_{i}(Y_i))$ where $m_{i}(Y_i)=\sum_{j \neq i} F_{ij}(Y_i,y_j)$. In a message passing interpretation, $m_{i}(\cdot) \in \bar{\mathbb{R}}^{|D^i|}$ represents the sum of all messages received from neighbor variables $Y_j$ through the incident functions $F_{ij}$, given $Y_j = y_j$. Computing the NPLL is in $O(n(n-1)d)$ per sample and epoch. It can easily be vectorized (computed independently on each variable).
\end{property}

The NPLL enables scalable training from natural inputs assuming an underlying GM NP-hard optimization problem. However, the proof of asymptotic consistency of the NPLL~\cite{besag1975statistical,geman1986markov} relies on identifiability assumptions that obviously do not hold in the context of constraints (zero probabilities) because of redundant constraints. Unsurprisingly, the NPLL is known to perform poorly in the presence of large costs~\cite{difficultGM}. Empirically, we observed that the resulting architecture completely fails at solving even the simplest symbolic Sudoku problem where $\omega$ contains the digits from the unsolved Sudoku grid and $\by$ is the corresponding solution.

\section{The E-NPLL}

To understand the incapacity of the NPLL to deal with large costs, it is interesting to look into the contribution of every pair $(\omega,\by)$ to the gradient $\frac{\partial NPLL}{\partial N(\omega)[i,j](v_i, v_j)}$ of the NPLL for a given pair of values $(v_i,v_j)$ of a pair of variables $(Y_i,Y_j)$.

\begin{property}[see full paper] The contribution of a pair $(\omega,\by)$ to the gradient $\frac{\partial NPLL}{\partial N(\omega)[i,j](v_i, v_j)}$ is equal to
  \begin{multline*}
    [\mathbb{1}(y_i=v_i, y_j=v_j) - P^{N(\omega)}(v_i | \by_{-i})\mathbb{1}(y_j=v_j)] \\ + [\mathbb{1}(y_i=v_i, y_j=v_j)  - P^{N(\omega)}(v_j | \by_{-j})\mathbb{1}(y_i=v_i)]
  \end{multline*}
\end{property}

The two terms in the gradient above come from NPLL terms computed on variables $Y_i$ and $Y_j$ respectively. Consider our previous example with four variables, $C=\{Y_1\neq Y_2, Y_2 + Y_3 > 1,Y_3\neq Y_4\}$ and $\by=(0,1,1,0)$. We focus on the variables $Y_{i=2}$ and $Y_{j=3}$ and assume that $C$ should hold under $\omega$, which means that the pair of values $(Y_2= 0, Y_3=0)$ should be predicted as forbidden. Being forbidden under $\omega$, $\mathbb{1}(y_2=0, y_3=0) = 0$.
If additionally, the forbidden pairs $(Y_1=0, Y_2=0)$ and $(Y_3=0, Y_4=0)$ have already reached a high cost under $\omega$, then both $P^{N(\omega)}(Y_2 = 0 | \by_{-2})$ and $P^{N(\omega)}(Y_3=0 | \by_{-3})$ will be close to zero, as well as the gradient itself. This will lead to a negligible (if any) change in the cost of the pair $(Y_2 = 0,Y_3 = 0)$: learning will be blocked or tremendously slowed down. Said otherwise, the fact that, in the context of $(\omega,\by)$, the forbidden pair $(Y_2=0, Y_3=0)$ is redundant w.r.t.\ already identified forbidden pairs $(Y_1=0,Y_2=0)$ and $(Y_3=0,Y_4=0)$ effectively prevents any change in the cost $N(\omega)[2,3](0,0)$. The issue with the NPLL lies therefore in the dynamic of the stochastic gradient optimization: the early identification of some high costs under $\omega$ will prevent the increase of other significant costs which are redundant in the context of the observed $\by$, but not redundant in the unconditioned problem.

Inspired by ``dropout'' in deep learning~\cite{srivastava2014dropout}, we introduce the Emmental NPLL (E-NPLL) as an alternative to the NPLL that should still work when constraints (infeasibilities) are present in $S$.

\begin{definition}\label{ENPLL} 
  The E-NPLL is a stochastic loss defined as
  $\textrm{E-NPLL}(\by)=-\sum_{Y_i\in\bY} \log(P^{N(\omega)}(y_i|\by_{-(\{i\}\cup\bM_i)}))$ where each $\bM_i$ is a random subset of $\{1, \ldots, n\}\setminus\{i\}$.
\end{definition}

In the above formula, $P^{N(\omega)}(y_i|\by_{-(\{i\}\cup\bM_i)})$ is a short (and slighly abusive) notation for  $\mathrm{softmax}(-m_{i}(Y_i))$ where $m_{i}(Y_i)=\sum_{j \not\in \{i\}\cup\bM_i} F_{ij}(Y_i,y_j)$. The idea of the E-NPLL follows directly from the previous gradient analysis: to prevent a combination of incident functions with already-learned high cost from shrinking gradients, we mute a random fraction of the incident functions. This is used in combination with an L1 regularization on the output of the learned network $N(\omega)$ to favour the prediction of exact zero costs which makes the GM optimization problem easier to solve. Because the E-NPLL is designed to fight the side effects of redundant constraints on gradients, we expect it to learn a GM $N(\omega)$ with all redundant pairwise constraints.

\subsection{Redundancy and Many Solutions}

We hypothesize that existing neural architectures where an exact solver is called during training will instead be insensitive to redundant constraints and will tend to not predict them. Contrarily to the NPLL that analyzes $N(\omega)$ independently for every variable $Y_i$ in the context of $\by_{-i}$, a solver will deal with the global problem and the presence or absence of globally redundant constraints will not affect its output and therefore the associated loss. With L1 regularization, we hypothesize that redundant constraints will tend to disappear. We will test this using the Hinge loss~\cite{tsochantaridis2005Hinge}, a well-known differentiable upper bound of the Hamming distance between a solver solution $\argmin_{\by\in D^{\bY}} N(\omega)(\by)$ and the observed $\by$. Note that in our settings, the Hinge loss is equivalent (under conditions detailed in the full paper) to the recent loss of~\cite{blackbox2}.

When a solver is called during training on $(\omega,\by)$, the output of the solver is compared to the solution $\by$. But reasoning problems usually have more than one solution and non-zero gradients get back-propagated when the predicted solution is different from the observed one, even if it is a correct one. It has been argued that this can severely hamper the training ability of solver-based layers, and Reinforcement Learning strategies have been defined to fight this~\cite{1ofMany}. Because it does not rely on the comparison of an arbitrary solver solution of the underlying discrete problem with the solution $\by$ in the data set, our approach should be directly able to deal with problems with many solutions.

\section{Related Works} 

As~\cite{RRN,satnet,OptNet,brouard2020pushing,blackbox,blackbox2}, we assume we have a data set of pairs $(\omega,\by)$ where $\by$ is sampled from a distribution of feasible high-quality solutions of a discrete reasoning problems whose parameters are influenced by $\omega$. We want to be able to predict solutions for new values of $\omega$. While~\cite{brouard2020pushing} proposed a non-differentiable architecture, most recent proposals, including ours, provide a differentiable DL architecture that enables learning from observables $\omega$ including natural inputs. All architectures exploit an optimization model with continuous parameters that are predicted by the previous layer. For training, a discrete differentiable reasoning layer can be used, requiring one~\cite{tsochantaridis2005Hinge,blackbox2,Berthet,iMLE} or two~\cite{blackbox} solver calls per training sample at each epoch. On NP-hard problems, such as MaxSAT or integer linear programming, these architectures may have excruciating training costs and are therefore applied only on tiny instances ($20$-cities traveling salesperson problems~\cite{blackbox} or 3 jobs/10 machines scheduling problems~\cite{TiasLTR}). For more scalability, relaxations of the underlying NP-hard problem using continuous optimization~\cite{OptNet,satnet,mandi2020interior,Wilder_Dilkina_Tambe_2019} or lifted message passing~\cite{RRN} have been used for efficient approximate solving.

For training, the architecture we propose relies instead on a dedicated loss function (that can therefore not be easily changed). At inference, the output of the architecture is a full GM that can be optimized with any suitable optimizer. This implies that our layer is the last layer of the architecture, which is probably a reasonable assumption for  most decision-focused learning problems~\cite{Wilder_Dilkina_Tambe_2019}.

In the \textit{Predict-and-optimize} framework~\cite{SPO,Mandi_SPO-relax}, a known optimization problem needs to be solved but some parameters $\bv$ in the criterion must be predicted using historical records of pairs $(\omega,\bv)$. $\bv$ being available at training, the optimization problem can be solved and the \emph{regret} (the loss in criteria generated by using predicted instead of true values of $\bv$) can be computed and used as a loss. In our case, we only have an unlabeled solution $\by$ instead of the optimization parameters $\bv$. The reader is referred to the review of~\cite{Kotary2021Review1} for more details on end-to-end constrained optimization learning.

Learning constraint is also central in constraint acquisition~\cite{bessiere2017constraint}, where guaranteed learning of constraints is achieved using positive/negative examples, in interaction with an agent. We instead assume a fixed data set with just positive examples annotated by features $\omega$, a setting where guarantees are essentially unreachable. Our approach is closer to~\cite{beldiceanu2016modelseeker}, where global (instead of elementary) constraints are learned from solutions of highly-structured problems, without differentiability.

\section{Experiments}

We test our architecture on logical (feasibility) problems with one or many solutions~\cite{1ofMany}, $\omega$ being purely symbolic or containing images. We also apply it to a real, purely data-defined, discrete optimization problem to check the ability of the E-NPLL to estimate a criteria.

Unless specified otherwise, all experiments use a Nvidia RTX-6000 with 24GB of VRAM and a $2.2$ GHz CPU with 128 GB of RAM. Our code is written in Python using PyTorch version 11.10.2 and PyToulbar2 version 0.0.0.2. We use the Adam optimizer with a weight decay of $10^{-4}$ and a learning rate of $10^{-3}$ (other parameters take default values). An L1 regularization with multiplier $2.10^{-4}$ is applied on the cost matrices $N(\omega)[i,j]$. Code and data are available at \url{https://forgemia.inra.fr/marianne.defresne/emmental-pll}.

\subsection{Learning To Play the Sudoku}

The NP-complete Sudoku problem is a classical logical reasoning problem that has been repeatedly used as a benchmark in a ``learning to reason'' context~\cite{RRN,OptNet,satnet,brouard2020pushing}. The task is to learn how to solve new Sudoku grids from a set of solved grids, without knowing the game rules.

\paragraph{Task.} Given samples $(\omega^l, \by^l)$ of initial and solved Sudoku grids, we want to learn how to solve new grids. Sudoku players know that Sudoku grids can be more or less challenging. As one could expect, it is also harder to train how to solve hard grids than easy grids~\cite{brouard2020pushing}. We use the number of initially filled cells (hints) as a proxy to the problem hardness, a grid with few hints being hard. The minimal number of hints required to define a single solution is 17, defining the hardest single-solution Sudoku grids. We use an existing data set~\cite{RRN}, composed of single-solution grids with $17$ to $34$ hints. We use $1,000$ grids for training, and $256$ for validation (all hardness). As in~\cite{RRN}, we test on the hardest 17-hints instances, $1,000$ in total.

A $9\times 9$ Sudoku grid is represented as $81$ cell coordinates with a possible hint when available. Each cell is represented by a GM variable with domain $\{1,\ldots,9\}$. For $N$, we use a Multi-Layer Perceptron (MLP) with $10$ hidden layers of $128$ neurons and residual connections~\cite{He_2016_CVPR} every $2$ layers. It receives the pairs of coordinates of pairs of cells $(Y_i,Y_j)$ and predicts all pairwise cost matrices $N(\omega)[i,j]$. Hints are used to set the values of their corresponding variable in $N(\omega)$. Performances are measured by the percentage of correctly filled grids; no partial credit is given for individual digits.

The resulting architecture learns how to solve the Sudoku and provides rules in $N(\omega)[i,j]$. For pairs of cells on the same row, column or $3\time 3$ sub-square, we expect soft difference-like cost function to be predicted (a matrix with a strictly positive diagonal and zeros) that prevents the use of the same value for the two variables. With the L1 regularization on the output of the neural net, other cost matrices should be $0$, indicating the absence of a pairwise constraint.

\paragraph{Grids with a unique solution.} We first train our network with the regular NPLL loss. As expected, it learns only a subset of the rules that suffices to make all other rules redundant: for a cell $Y_i$ in the context of $\by_{-i}$, a single clique of difference constraints for every row (or column, or square) is sufficient to determine the value of the cell $Y_i$ from $\by_{-i}$, creating vanishing gradients for all other constraints that are instead estimated as constant $0$ matrices. On the test set, inference completely fails.

We replaced the NPLL by the E-NPLL, ignoring messages from $k$ randomly chosen other variables. In terms of accuracy, the training is largely insensitive to the value of the hyper-parameter $k$ (see Table~\ref{table_k}) as long as it is neither $0$ (regular NPLL) nor close to $n-1$ (no information). However, larger values of $k$ tend to lead to longer training. We set $k = 10$ for all Sudoku experiments. In this case, training takes less than $15$ minutes. At inference, the predicted $N(\omega)$ leads to 100\%\ accuracy on the $1,000$ hard test-set grids.

\begin{table}[htb]
  \centering
  \begin{tabular}{lllp{25mm}}
    \toprule
    $k$  & Epochs         & Training time (s) & Runs with 100\% test grids solved \\
    \midrule
    $0$  & $100$          & -                 & $0$\%                             \\
    $10$ & $23.2 \pm 2.6$ & $566 \pm 67$      & $100$\%                           \\
    $20$ & $38.6 \pm 6.9$ & $900 \pm 151$     & $90$\%                            \\
    $50$ & $50.4 \pm 7.6$ & $1257 \pm 184$    & $90$\%                            \\
    $70$ & $27.2 \pm 2.7$ & $724 \pm 83$      & $100$\%                           \\
    $80$ & $100$          & -                 & $0$\%                             \\
    \bottomrule
  \end{tabular}
  \caption{Average performances over $10$ initialization, for various number of E-NPLL holes $k$\label{table_k}. Training is limited to $100$ epochs.}
\end{table}

In Table~\ref{table_Hinge}, we compare our results with previous approaches that learn how to solve Sudoku. The Graph Neural Net approach of Recurrent Relational Network (RRN)~\cite{RRN}, the convex optimization layer SATNet~\cite{satnet} and a hybrid ML/CP method~\cite{brouard2020pushing}. It should be noted that SATNet's accuracy is measured on a test set of easy Sudoku grids (avg.~36.2 hints). All the methods that rely on an exact solver are able to reach 100\% accuracy but the modeling flexibility of Deep Learning offers the ability to describe the grid geometry in $\omega$, leading to better inductive bias and far better data efficiency. In our case, with $k=10$, we still obtain 100\% accuracy on 17-hints grids using a training set of just $200$ grids. More epochs are necessary (less than $200$) but the training time does not increase ($587 \pm 32s$ over 10 training using 10 different seeds).

\begin{table}[hb]
  \centering\setlength{\tabcolsep}{3.5pt}
  \begin{tabular}{rlllr}
    \toprule
    Approach                  & Acc.           & \#hints & Train set    & Param. \\\midrule
    \cite{RRN}                & ~96.6\%        & 17      & 180,000      & 200k   \\
    \cite{satnet}             & ~99.8\%        & 36.2    & 9,000        & 600k   \\
    \cite{brouard2020pushing} & \textbf{100\%} & 17      & 9,000        & -      \\
    Hinge (here)              & \textbf{100\%} & 17      & 1,000        & 180k   \\
    \textbf{E-NPLL} (here)    & \textbf{100\%} & 17      & \textbf{200} & 180k   \\
    \bottomrule
  \end{tabular}
  \caption{Accuracies of related works. The '\# hints' gives the hardness of the test set. Param. is the number of parameters of the nets.\label{table_Hinge}}
\end{table}

\paragraph{Interpreting the neural output.} In our settings, besides a solution, we obtain a full GM that can be scrutinized and interpreted in terms of which rules have been learned. In the benchmarking situation of the Sudoku problem, we can be confident that the accuracy of 100\%\ observed on the test set actually extends to any Sudoku instance. More interestingly, as shown in~\cite{lim2022learning}, this output could also be monitored during training to detect symmetries that could speed up training and provide more human-readable information.

\begin{figure}
  \centering
  \includegraphics[width=\linewidth]{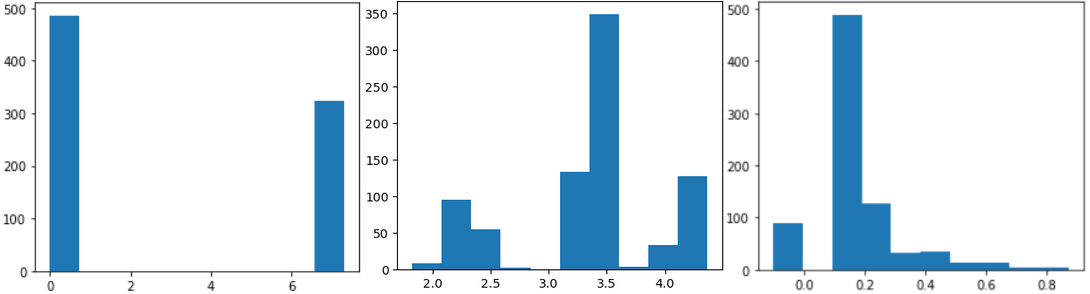}
  \caption{Diagonal costs learned for pairs of cells  on the same row, column or sub-square, using the regular NPLL (left), the E-NPLL with $k=10$ (middle), and the Hinge loss (right). A positive cost shows a constraint has been learned. The NPLL shows many missed constraints. The E-NPLL learns all $810$ constraints, the Hinge loss omits some redundant ones (preserving $100\%$ test accuracy).}
  \label{fig:hist}
\end{figure}

\paragraph{Redundant constraints.} The rules of Sudoku are naturally described as $810$ pairwise difference constraints. It is known that $162$ of these (at least) are redundant~\cite{demoen2014redundant}. As we expected, the E-NPLL produces the full set of $810$ pairwise rules, including redundant constraints. We tried to learn how to solve the Sudoku problem using an embedded exact prover and the Hinge loss~\cite{tsochantaridis2005Hinge}. Embedding an exact prover for an NP-hard problem as a neural layer proved to be challenging even on this small problem. In particular, in the first epochs, the predicted GMs $N(\omega)$ are dense random pairwise GMs that are impossible to solve in a reasonable time~\cite{zhang2001phase}. Therefore, we adopted a progressive learning scheme where additional hints are extracted from $\by$, leaving an increasing number of unassigned variables: we used $20$ variables initially, increasing this at each epoch until, eventually, only the initial hints were left. This schedule was difficult to adjust. Each training took 2 to 3 days on $1,000$ training grids. Still, it reaches 100\% accuracy on the Sudoku test set. Depending on initialization, we observe that several redundant constraints are not learned (see Figure~\ref{fig:hist}), always preserving more than $648$ soft difference cost matrices, the conjectured number of non-redundant constraints for Sudoku~\cite{demoen2014redundant}.

\paragraph{Multiple-solution grids.} Published Sudoku grids have only one solution. We now consider a Sudoku benchmark used in~\cite{1ofMany}, where each grid has more than one solution. For each grid, the set of solutions is only partially accessible during training: at most $5$ of them are present in the training set. The aim is to be able to predict any solution. We use $1,000$, $64$ and $256$ grids of the data set from~\cite{1ofMany} respectively for training, validating, and testing. All hyper-parameters are set as previously, the validation set is only used to stop training. The testing criteria is to be able to retrieve one of the feasible solutions (all of them are known for testing).

Since the E-NPLL never compares a solver-produced solution to the provided solution $\by$, it is not sensitive to the existence of many solutions. With the same training procedure as previously, selecting one of the $5$ provided solutions randomly at each epoch, training takes $723.4 \pm 64.9$ seconds ($21.4 \pm 1.9$ epochs) on average, leading to one of the expected solutions for 100\% of the test grids. In fact, since the correct rules are identified, we could verify that thresholding the learned costs into Boolean enables a complete enumeration of all feasible solutions for all instances in the test set.

\subsection{Visual Sudoku}

Our previous examples show the benefits of exploiting inputs $\omega$. To explore this capacity more deeply, we tackle the visual Sudoku problem in which hints are MNIST images. The goal is to simultaneously learn how to recognize digits and how to play Sudoku. We add a LeNet network~\cite{LeNet} to our previous architecture. The GM $N(\omega$) produced is composed of the pairwise cost functions produced by the same Residual MLP with the addition of the negated logit output of LeNet as an elementary cost function on each variable $Y_i$ with a hint. This GM is fed to our regularized E-NPLL loss for back-propagating solutions. No new tuning is necessary (same learning rate for both networks).

Our data set is obtained from the symbolic Sudoku data set by replacing hints with corresponding MNIST images, as in~\cite{brouard2020pushing}. We use the same $1,000$ grids for training, the validation set contains $64$ grids (only used to stop training). To check for sensitivity to initialization (an 80\% training failure rate was observed for SATNet in~\cite{assessingSATNet}), $10$ runs with different seeds were performed.

After training, we extract the LeNet network alone: it reaches a $97.6$\% accuracy on MNIST, being indirectly supervised by the provided hints, through the E-NPLL (directly in SATNet~\cite{assessingSATNet}). We test again on $1,000$ hard grids (see Table~\ref{table_visual}). When all the hints are correctly predicted, grids are correctly filled, as proper rules have been learned. Moreover, in $8.7$\% of cases, the solver is able to correct LeNet's errors, leading to an overall accuracy of $76$\% on hard grids. Training takes an average time of $1150 \pm 13 s$.

\begin{table}[htb]
  \centering
  \begin{tabular}{ccc}
    \toprule
    MNIST accuracy   & Correct grids & Of which corrected \\
    \midrule
    $97.6 \pm 0.9$\% & $760 \pm 9$   & $87 \pm 3$         \\
    \bottomrule
  \end{tabular}
  \caption{Visual Sudoku performance ($1,000$ hard 17-hints test grids)\label{table_visual}}
\end{table}

We also compare our architecture with SATNet, using their data set of 9K training and 1K easy test grids (average of $36.2$ hints~\cite{satnet}). We use the exact same parameters as in the previous experiment and reuse SATNet's ConvNet to process MNIST digits. We train for at most $20$ epochs ($100$ for SATNet) using $64$ of the training grids for validation (to decide when to stop training). On this data set, SATNet's accuracy is $63.2$\%. SATNet's authors compared this to a theoretical maximum accuracy of $74.7\%$, using a $99.2\%$ accuracy MNIST classifier and a perfect Sudoku solver. Integrating LeNet's uncertainty on classification as negated logits in the GM pushes accuracy well beyond this theoretical limit (see Table~\ref{table_visual2}). Since the ConvNet is trained through the E-NPLL loss, its weights are automatically adjusted to optimize the joint Pseudo-Loglikelihood that includes also the Sudoku rules being learned. This automatically calibrates its output for the task (similarly to what is done, \textit{a posteriori} and with known hard Sudoku rules, in~\cite{mulamba2020hybrid}).

\begin{table}[h]
  \centering
  \begin{tabular}{ccc}
    \toprule
    SATNet    & Theoretical & Ours                      \\ 
    \midrule
    $63.2$ \% & $74.2$\%    & $\mathbf{94.1} \pm 0.8$\% \\ 
    \bottomrule
  \end{tabular}
  \caption{Fraction of solved grids using SATNet data set for training and testing (averaged over 3 different initializations).\label{table_visual2}}
\end{table}

\subsection{Learning To Design Proteins}

The problem of designing proteins has similarities with solving Sudoku~\cite{STROKACH2020ProteinSolver}. Proteins are linear macro-molecules defined by a sequence of natural amino acids. Proteins usually fold into a specific 3D structure which enables specific biological or biochemical functions. Designed proteins have applications in health and green chemistry, among others~\cite{kuhlman2019advances}. To design new proteins, with new functions or enhanced properties, one usually starts from an input backbone structure, matching the target functions, and predicts a sequence of amino acids that will fold onto the target structure.

Considering an input protein structure as a Sudoku grid, each amino acid corresponds to a cell and must be chosen among the $20$ natural amino acids instead of $9$ digits. The structure is predominantly determined by inter-atomic forces, which drive folding into a minimum energy geometry: the most usual approach of the protein design problem is as an NP-hard unconstrained energy minimization problem~\cite{pierce2002protein,allouche2014computational}. Inter-atomic forces are influenced by relative distances and atomic natures which implies that the final interactions inside a protein depend on the geometry of the input structure: we will therefore use this geometry in the natural inputs $\omega$, as we did with the (fixed) Sudoku grid geometry. Compared to Sudoku, protein design instances have variable geometry, they may contain several hundreds of amino acids or more and they are subjected to many-bodies interactions that cannot be directly represented in a pairwise GM.

When designing proteins, the Hamming distance between the predicted and observed (native) sequences, called the Native Sequence Recovery rate (NSR), is often used for evaluation. Protein design is a multiple-solution problem: above 30\% of similarity, two protein sequences are considered as having the same fold (geometry). So, one given structure can adopt many sequences and a 100\%\ NSR cannot be reached.

For training, we use the data set of~\cite{Ingraham2019}, already split into train/validation/test sets of respectively $17,000$, $600$ and $1,200$ proteins, in such a way that proteins with similar structures or sequences are in the same set. Similarly to Sudoku, a protein is described by features $\omega$ computed on each pair of positions $i,j$ in the protein. They include inter-atomic distance features encoded with Gaussian radial basis function~\cite{ProteinMPNN}, and a positional encoding of the sequence distance ($|i-j|$)~\cite{Vaswani2017}. Each backbone geometry $\omega$ is associated with $\by$, the sequence of the corresponding known protein. All pair features in $\omega$ are processed by a neural network composed of a gated MLP~\cite{liu2021pay} that learns an environment embedding from a central amino acid and its neighbors within $10$\AA, fed to a ResMLP (as for Sudoku) that takes pairs of features and environment embeddings to predict $20\times 20$ cost matrices. Training and architecture details are in annex~\ref{annex:archi}.

We train the same model with the same initialization using either the NPLL or the E-NPLL loss. To adapt to variable protein sizes, the E-NPLL eliminates $k$\% of incoming neighbor messages. With up to $500$ amino acids, the optimization task is challenging at inference and we used a recent GM convex solver~\cite{durante:hal-03673535}. As shown in Table~\ref{table_protein}, we observe that the E-NPLL not only preserves the good properties of the NPLL but actually improves the NSR. While protein design is often stated as an unconstrained optimization problem, we hypothesize that this improvement results from the existence of infeasibilities: when local environments are very tight, they absolutely forbid large amino acids. Such infeasibilities could not be properly estimated by the NPLL alone.

\begin{table}[h]
  \centering
  \begin{tabular}{cccc}
    \toprule
    k   & 0      & 10     & 40              \\
    \midrule
    NSR & 40.6\% & 41.4\% & \textbf{42.9\%} \\
    \bottomrule
  \end{tabular}
  \caption{Comparing the E-NPLL and the regular NPLL on the test proteins. Median NSR over the full test set are given. The case $k=0$ corresponds to the regular NPLL. \label{table_protein}}
\end{table}

Our architecture provides a decomposable scoring function, such as those used for protein design in Rosetta~\cite{RosettaDecoy}.  Table~\ref{table_Rosetta} compares both approaches on the data set of~\cite{RosettaDecoy}. We see that the E-NPLL learned decomposable scoring function outperforms Rosetta's energy function. This is all the more satisfying as Rosetta's full-atom function considers all atoms of the protein while we just use the backbone geometry and amino acids identities (as in coarse-grained scoring functions~\cite{coarse-grain}).

\begin{table}[htb]
  \centering
  \begin{tabular}{ccc}
    \cmidrule[\heavyrulewidth]{2-3}
        & Rosetta\footnotemark & E-NPLL          \\
    \midrule
    NSR & 17.9\%               & \textbf{27.8\%} \\
    \bottomrule
  \end{tabular}
  \caption{Comparison with the energy-based design method Rosetta\label{table_Rosetta} on small single-chain proteins.}
\end{table}
\footnotetext{Rosetta's results are extracted from~\cite{Ingraham2019}}

\section{Conclusion}

In this paper, we introduce a hybrid neural+graphical model architecture and a dedicated loss function for learning how to solve discrete reasoning problems. It is differentiable and as such, allows natural inputs to participate in the definition of discrete reasoning/optimization problems, providing the ability to inject suitable inductive biases that can also enhance data efficiency, as shown in the case of Sudoku. While most discrete/relaxed optimization layers~\cite{blackbox,blackbox2,satnet} can be inserted in an arbitrary position in a neural net, our \textit{final} GM layer with the E-NPLL loss offers scalable training, avoiding calls to exact solvers that quickly struggle with the noisy instances that are predicted in early training epochs. It is able to benefit from exact or relaxed solvers during inference. Thanks to the E-NPLL, it can simultaneously identify a criterion to optimize as well as constraints. Finally, its output can be scrutinized to check properties and can be \emph{a posteriori} completed with side constraints or additional criteria, in order to inject further instance-dependent information, that may have been learned, be available as knowledge or as user requirements.

On various NP-hard Sudoku benchmarks, it is able to produce correct solutions from natural input (including images), while being data-efficient and capable of generalization in incomplete multiple-solution settings. The use of an exact prover during inference results in robust prediction, allowing for the correction of otherwise noisy neural predictions. On the real-world protein design problem, our approach is also quickly able to learn a geometry-dependent pairwise decomposed function that outperforms the most recent full-atom pairwise decomposable energy functions for predicting the sequence of natural proteins.

Much remains to be done around this architecture. As for SATNet~\cite{lim2022learning}, the ultimate $N(\omega)$ GM layer of our architecture could be analyzed during training to identify emerging hypothetical global properties such as symmetries or global decomposable constraints, allowing for more efficient learning and improved human understanding. For memory and computational efficiency, we limited ourselves to pairwise models but the use of other languages (e.g. weighted clauses) in replacement of, or addition to, pairwise functions would enhance the capacity of the architecture to capture many-bodies interactions. Another possibility is the use of latent/hidden variables~\cite{stergiou1999encodings}.

\appendix

\section{On the Hinge loss}

\begin{property}
  The Hinge loss for the Hamming loss with a $0$-margin is equivalent to the loss of~\cite{blackbox2} with no projection.
\end{property}

When using the Hinge loss, the solver is embedded as a neural layer. For each sample $(\omega,\by)$, the solver computes the optimal solution $\by^*$ of the predicted GM $N(\omega)$: $\by^* = \argmin_{\bt\in D^{\bY}} N(\omega)(\bt)$.

The observed  $\by$ and the predicted solution $\by^*$ are compared using the Hamming loss.

$$\textit{Hamming}(\by, \by^*) = \sum_{i=1}^{n} \mathbb{1}[{y_i \neq y^*_i}]$$

This loss is discrete, therefore its gradient are uninformative (either $0$ or non-existent). To obtain meaningful gradients to back-propagate, the Hinge loss~\cite{tsochantaridis2005Hinge} provides an upper bound on the loss that is specifically easy to express for pairwise-decomposable losses $L$, such as the Hamming loss above.

\begin{align*}
  \textit{Hinge}(\omega, \by) & = \max_{\bt\in D^\bY} \left[ L(\by,\bt) + (N(\omega)(\by) - N(\omega)(\bt)) \right]                          \\
                              & = N(\omega)(\by) -\underbrace{\min_{\bt\in D^\bY}\left[N(\omega)(\bt) - L(\by,\bt)\right ]}_{\argmin= \by^m}
\end{align*}
The Hamming loss can be simply computed as the cost of a CFN over $\bY$ with one cost function $F_i$ per variable $Y_i$ with $F_i(t_i) = \mathbb{1}[t_i \neq y_i]$. The right minimization problem above can be therefore solved by calling the GM solver on the original problem $N(\omega)$ to which functions $-F_i$ scaled by factor $\alpha\geq 0$ (called the margin) have been added. If we denote by $\by^m$ the solution of this problem, for sample $(\omega,\by)$, the Hinge loss will contain costs N$(\omega)[i,j](v_i,v_j)$ with a positive sign iff $v_i=y_i$ and $v_j=y_j$ (in $N(\omega)(\by)$) and also with a negative sign iff $v_i=y^m_i$ and $v_j=y^m_j$ ($-N(\omega)(\by^m)$). The only possibly non-zero gradient terms will be therefore $+1$ for $N(\omega)[i,j](y_i,y_j)$ and $-1$ for $N(\omega)[i,j](y^m_i,y^m_j)$ which will cancel iff $y_i=y^m_i$ and $y_j=y^m_j$.
When looking at equation (2) from~\cite{blackbox2}, we obtain exactly the same gradients using $\alpha = 0$. Therefore, the $0$-margin Hinge loss (also called the contrastive Viterbi loss or the perceptron loss~\cite{lecun2006tutorial}) is equivalent to the loss of ~\cite{blackbox2} with no projection.

\section{Gradient of the NPLL}

We have a data set $S$ composed of $m$ pairs $(\omega^l,\by^l), 1\leq l\leq m$. The Negative Pseudologlikelihood of $S$ is the sum of the negative log-probability of each $(\omega^l,\by^l)$:

\begin{align*}
  NPLL(S) & = \sum_{l=1}^m NPLL(\omega^l,\by^l)                                                 \\
          & = -\sum_{l=1}^m \left[\sum_{Y_i\in \bY} \log P^{N(\omega)}(y_i^l|\by^l_{-i})\right]
\end{align*}

where
$$P^{N(\omega)}(y_i^l|\by^l_{-i}) =
  \frac{\exp(-\sum_{j\neq i} N(\omega)[i,j](y^l_i,y^l_j))}{\sum_{v_i\in D^i} \exp(-\sum_{j\neq i} N(\omega)[i,j](v_i,y^l_j))}$$

The conditional probability above is obtained using the normalizing constant $Z^{N(\omega)}(\by^l_{-i})$ in the  denominator:

$$Z^{N(\omega)}(\by^l_{-i}) = \sum_{v_i\in D^i} \exp(-\sum_{j\neq i} N(\omega)[i,j](v_i,y^l_j))$$

\noindent computed over all possible values $v_i$ of $Y_i$. This corresponds to the application of a softmax on the logits $-\sum_{j\neq i} N(\omega)[i,j](v_i,y^l_j)$.

Minimizing the NPLL means maximizing the probability above, therefore making  $-N(\omega)[i,j](\cdot,y^l_j)$ higher on the observed value $y^l_i$ (used in the numerator) than on the other values $v_i\neq y^l_i$ or equivalently, the cost $N(\omega)[i,j](\cdot,y^l_j)$ lower on $y^l_i$ than on other values: the NPLL is a contrastive loss that seeks to create a margin between the values that are observed in the sample $S$ and the other values of the variable, for every variable and every sample.

Focusing on one pair $(\omega,\by)\in S$ , we expand and get:
\begin{multline}\label{NPLL1}
  NPLL(\omega,\by) =\\-\sum_{Y_i\in \bY} \left[ \left(-\sum_{j\neq i} N(\omega)[i,j](y_i,y_j)\right) - \log Z^{N(\omega)}(\by_{-i})\right]
\end{multline}

The NPLL is a sum over all variables $Y_i\in \bY$ and we consider the contribution of a given variable $Y_i$. To compute the gradients of the corresponding term of the NPLL, we first compute the partial derivative of the logarithm of the normalizing constant $Z^{N(\omega)}(\by_{-i})$ ($i$ fixed) w.r.t.\ $N(\omega)[i,j](v_i,y_j)$ (for arbitrary $j\neq i$ and $v_i\in D^i$, other costs do not appear in $Z^{N(\omega)}(\by_{-i})$  and the corresponding partial derivative is $0$).

\begin{align*}
  \frac{\partial \log Z^{N(\omega)}(\by_{-i})}{\partial N(\omega)[i,j](v_i,y_j)} & = \frac{-\exp(-\sum_{k\neq i} N(\omega)[i,k](v_i,y_k))}{Z^{N(\omega)}(\by_{-i})} \\
                                                                                 & = -P^{N(\omega)}(v_i|\by_{-i})
\end{align*}

For any $Y_i$, the partial derivative of the first term in equation~\ref{NPLL1} w.r.t. $N(\omega)[i,j](v_i,v_j)$ is $-\mathbb{1}(v_i=y_i,v_j=y_j)$.

Overall, given that $ N(\omega)[i,j](v_i, v_j)$ and $N(\omega)[j,i](v_j, v_i)$ are the same, the contribution of sample $(\omega,\by)$ to $\frac{\partial NPLL}{\partial N(\omega)[i,j](v_i, v_j)}$ will reduce to the non-zero contributions of variables $Y_i$ and $Y_j$:
\begin{multline*}
  \frac{\partial NPLL}{\partial N(\omega)[i,j](v_i, v_j)} = \\ [\mathbb{1}(y_i=v_i, y_j=v_j) - P^{N(\omega)}(v_i | \by_{-i})\mathbb{1}(y_j=v_j)] \\ + [\mathbb{1}(y_i=v_i, y_j=v_j)  - P^{N(\omega)}(v_j | \by_{-j})\mathbb{1}(y_i=v_i)]
\end{multline*}

\section{Architecture and training details}
\label{annex:archi}

For the protein design task, the neural network is composed of an input linear layer of size $128$, a gatedMLP with 15 layers, a width of $128*6=768$ and an output dimension of $64$, and a ResMLP with $20$ layers of $256$ neurons, with residual connections every $2$ layers. The gatedMLP consider the $45$ nearest neighbours of each residues. As in most protein energy functions~\cite{alford2017rosetta}, amino acid pairs separated by a large distance are ignored. We use a $15$\AA\ threshold.

The neural network is trained using the Adam optimizer, with a weight decay of $10^{-3}$ and an initial learning rate of $5.10^{-4}$, divided by $10$ when the validation loss decreases (with patience $0$) until it reaches $10^{-8}$. A L1 regularization of $10^{-4}$ is applied to costs.

To compare between the NPLL loss and the E-NPLL loss, we trained the same model with the same hyperparameters and starting from the same weight initialization with each of the loss. Both models run as long as the minimum LR is not reached, and therefore not necessarily for the same number of epochs.

\section*{Acknowledgements}
This work was performed using HPC resources from CALMIP (Grant 2022-P21025), and Jean-Zay GENCI-IDRIS (Grant 2022-AD011013779) and has been supported by the French ANR through grant ANR-19-PIA3-0004 (ANITI). M. Defresne PhD studies are funded by the EUR BioEco (grant ANR-18-EURE-0021). We thank B. Savchynskyy for referring us to the Hinge loss, and the proof-reading team from MIAT. We thank Romain Gambardella for detecting issues in the final NPLL gradient formula and suggestions for disambiguation of Definition~\ref{ENPLL}.

\end{document}